\documentclass{article}

\PassOptionsToPackage{numbers, compress}{natbib}
\usepackage[final,nonatbib]{nips_2016}

\usepackage[utf8]{inputenc} 
\usepackage[T1]{fontenc}    
\usepackage{hyperref}       
\usepackage{url}            
\usepackage{booktabs}       
\usepackage{amsfonts}       
\usepackage{nicefrac}       
\usepackage{microtype}      
\usepackage[pdftex]{graphicx}
\usepackage{amsmath,amssymb}

\usepackage{bm}
\newcommand{\vect}[1]{\boldsymbol{\mathbf{#1}}}
\usepackage{color,soul}

\DeclareMathOperator{\E}{\mathbb{E}}

\title{Learning Online Alignments with Continuous Rewards Policy Gradient}

\author{
	Yuping Luo\thanks{Work done as intern at Google Brain}\\
  Institute for Interdisciplinary Information Sciences\\
  Tsinghua University\\
  \texttt{roosephu@gmail.com} \\
  \And
  Chung-Cheng Chiu \\
  Google Brain \\
  \texttt{chungchengc@google.com} \\
  \AND
  Navdeep Jaitly \\
  Google Brain \\
  \texttt{ndjaitly@google.com} \\
  \And
  Ilya Sutskever \\
  Open AI\thanks{Work done at Google Brain} \\
  \texttt{ilyasu@openai.com} \\
}

\begin{document}

\maketitle
\begin{abstract}
Sequence-to-sequence models with soft attention had significant success
in machine translation, speech recognition, and question answering.  Though
capable and easy to use, they require that the entirety of the
input sequence is available at the beginning of inference, an assumption that is not valid
for instantaneous translation and speech recognition.  To address this problem,
we present a new method for solving
sequence-to-sequence problems using hard online alignments instead of
soft offline alignments.  The online alignments model is able to start
producing outputs without the need to first process the entire input
sequence.  A highly accurate online sequence-to-sequence model is
useful because it can be used to build an accurate voice-based
instantaneous translator. Our model uses hard binary stochastic decisions
to select the timesteps at which outputs will be produced. The model is trained
to produce these stochastic decisions using a standard policy gradient
method.  In our experiments, we show that this model achieves encouraging
performance on TIMIT and Wall Street Journal (WSJ) speech
recognition datasets.
\end{abstract}

\section{Introduction}

Sequence-to-sequence models \cite{sutskever-nips-2014,cho-emnlp-2014}
are a general model family for solving supervised learning problems
where both the inputs and the outputs are sequences.  The performance
of the original sequence-to-sequence model has been greatly improved
by the invention of \emph{soft attention} \cite{bahdanau-iclr-2015},
which made it possible for sequence-to-sequence models to generalize
better and achieve excellent results using much smaller networks on
long sequences.  The sequence-to-sequence model with attention had
considerable empirical success on machine translation
\cite{bahdanau-iclr-2015}, speech recognition
\cite{chorowski-nips-2014,chan2015listen}, image caption generation
\cite{xu-icml-2015,vinyals-arvix-2014}, and question answering
\cite{weston2014memory}.

Although remarkably successful, the sequence-to-sequence model with
attention must process the entire input sequence before producing an
output.  However, there are tasks where it is useful to start
producing outputs before the entire input is processed. These tasks
include both speech recognition and machine translation, especially
because a good online speech recognition system and a good online
translation system can be combined to produce a voice-based
instantaneous translator (also known as a Babel Fish
\cite{adams1995hitch}), which is an important application.

In this work, we present a simple online sequence-to-sequence model
that uses binary stochastic variables to select the timesteps at which
to produce outputs. The stochastic variables are trained with a policy
gradient method (similarly to Mnih et al.~\cite{mnih2014recurrent} and
Zaremba and Sutskever \cite{zaremba2015reinforcement}).  Despite its
simplicity, this method achieves encouraging results on the TIMIT and
the Wall Street Journal speech recognition datasets. Our results
suggest that a larger scale version of the model will likely achieve
state-of-the-art results on many sequence-to-sequence problems.

\subsection{Relation To Prior Work}

While the idea of soft attention as it is currently understood was
first introduced by Graves \cite{graves2013generating}, the first
truly successful formulation of soft attention is due to Bahdanau et
al.~\cite{bahdanau-iclr-2015}. It used a neural architecture that
implements a ``search query'' that finds the most relevant element in
the input, which it then picks out.  Soft attention has quickly become
the method of choice in various settings because it is easy to implement
and it has led to state of the art results on various tasks. For example,
the Neural Turing Machine \cite{graves2014neural} and the Memory Network
\cite{sukhbaatar2015end} both use an attention mechanism similar to
that of Bahdanau et al.~\cite{bahdanau-iclr-2015} to implement models
for learning algorithms and for question answering.

While soft attention is immensely flexible and easy to use, it assumes
that the test sequence is provided in its entirety at test time.  It
is an inconvenient assumption whenever we wish to produce the relevant
output as soon as possible, without processing the input sequence in
its entirety first. Doing so is useful in the context of a speech
recognition system that runs on a smartphone, and it is especially
useful in a combined speech recognition and a machine translation
system.

There exists prior work that investigated methods for producing an
output without consuming the input in its entirety.  These include the
work by Mnih \cite{mnih2014recurrent} and Zaremba and Sutskever
\cite{zaremba2015reinforcement} who used the Reinforce algorithm to
learn the location in which to consume the input and when to emit an
output.  Finally, Jaitly et al.~\cite{jaitly2015online} used an
online sequence-to-sequence method with conditioning on partial
inputs,  which yielded encouraging results on the TIMIT dataset.
Our work is most similar to Zaremba and Sutskever \cite{zaremba2015reinforcement}.
However, we are able to simplify the learning problem for
the policy gradient component of the algorithm by using only
one stochastic decision per time step, which makes the model much more
effective in practice.

\section{Methods}
\begin{figure}[t]
  \centering
  \includegraphics[width=0.8\linewidth]{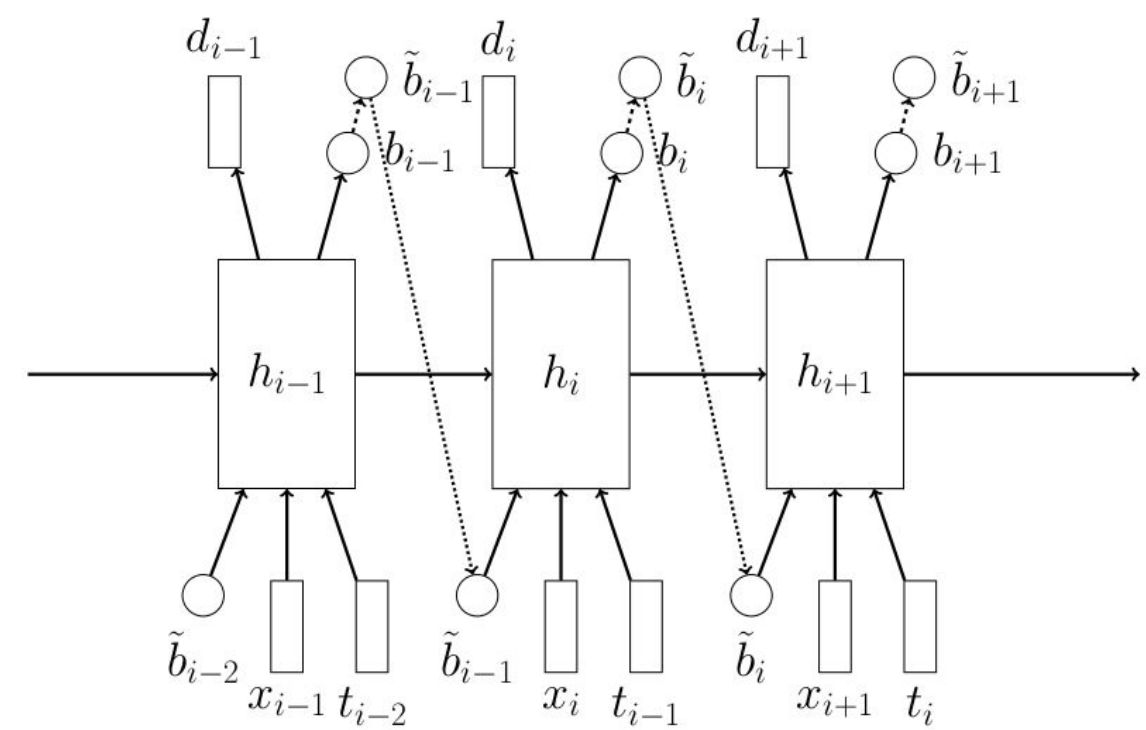}
  \caption{Overall Architecture of the model used in this paper.}\label{fig:model}
\end{figure}

In this section we describe the details of our recurrent neural
network architecture, the reward function, and the training and
inference procedure. We refer the reader to figure~\ref{fig:model} for
the details of the model.

We begin by describing the probabilistic model we used in this
work. 
At each time step, $i$, a recurrent neural network (represented in figure 1) decides whether to emit an output token. The decision is made by a stochastic binary logistic unit $b_i$. Let $\tilde{b}_i \sim \text{Bernoulli}(b_i)$ be a Bernoulli distribution such that if $\tilde{b}_i$ is 1, then the model outputs the vector $d_i$, a softmax distribution over the set of possible tokens. The current position in the output sequence $y$ can be written $\tilde{p}_i = \sum_{j=1}^i \tilde{b}_j$, which is incremented by 1 every time the model chooses to emit. Then the model's goal is to predict the desired output $y_{\tilde{p}_i}$; thus whenever $\tilde{b}_i = 1$, the model experiences a loss given by
\[\text{softmax\_logprob}(d_i;y_{\tilde{p}_i}) = - \sum_{k}\log(d_{ik})y_{\tilde{p}_i k}\]
where $k$ ranges over the number of possible output tokens.


At each step of the RNN, the binary decision of the previous timestep,
$\tilde{b}_{i-1}$ and the previous target $t_{i-1}$ are fed into the
model as input.  This feedback ensures that the model's outputs are maximally
dependent and thus the model is from the sequence to sequence family.

We train this model by estimating the gradient of the log probability
of the target sequence with respect to the parameters of the
model. While this model is not fully differentiable because it uses
non-diffentiable binary stochastic units, we can estimate the
gradients with respect to model parameters by using a policy gradient
method, which has been discussed in detail by Schulman et
al.~\cite{schulman2015gradient} and used by Zaremba and Sutskever
\cite{zaremba2015reinforcement}.

In more detail, we use supervised learning to train the network to
make the correct output predictions, and reinforcement learning to
train the network to decide on when to emit the various
outputs.  Let us assume that the input sequence is given by $(x_1,\ldots,x_{T_1})$
and let the desired sequence be $(y_1,\ldots,y_{T_2})$, where $y_{T_2}$ is a special
end-of-sequence token, and where we assume that $T_2 \leq T_1$. Then the log
probability of the model is given by the following equations:
\begin{eqnarray}
  h_i &=& \textrm{LSTM}(h_{i-1}, \mathrm{concat}(x_i, \tilde{b}_{i-1}, \tilde{y}_{i-1})) \\
  b_i &=& \mathrm{sigmoid}(W_\mathrm{b} \cdot h_i) \\
  \tilde{b}_i &\sim & \mathrm{Bernoulli}(b_i) \\
  \tilde{p}_i &=& \sum_{j=1}^i \tilde{b}_j \\
  \tilde{y}_i &=& y_{\tilde{p}_i} \\
	d_i &=& \mathrm{softmax}(W_o h_i) \\
  \mathcal{R}  &=& \mathcal{R} + \tilde{b}_i \cdot \text{softmax\_logprob}(d_i; \tilde{y}_i)  \label{eqn:reward}
\end{eqnarray}

In the above equations, $\tilde{p}_i$ is the ``position'' of the model
in the output, which is always equal to $\sum_{k=1}^i \tilde{b}_i$:
the position advances if and only if the model makes a prediction.
Note that we define $y_0$ to be a special beginning-of-sequence
symbol.  The above equations also suggest that our model can easily be implemented
within a static graph in a neural net library such as TensorFlow, even though
the model has, conceptually, a dynamic neural network architecture.

Following Zaremba and Sutskever~\cite{zaremba2015reinforcement}, we modify the model from
the above equations by forcing $\tilde{b}_i$ to be equal to 1
whenever $T_1 - i \leq T_2 -\tilde{p}_i$.  Doing so ensures that
the model will be forced to predict the entire target sequence $(y_1,\ldots,y_{T_2})$,
and that it will not be able to learn the degenerate solution where
it chooses to never make any prediction and therefore never experience
any prediction error.

\begin{figure*}[h]
  \centering
  \includegraphics[width=\linewidth]{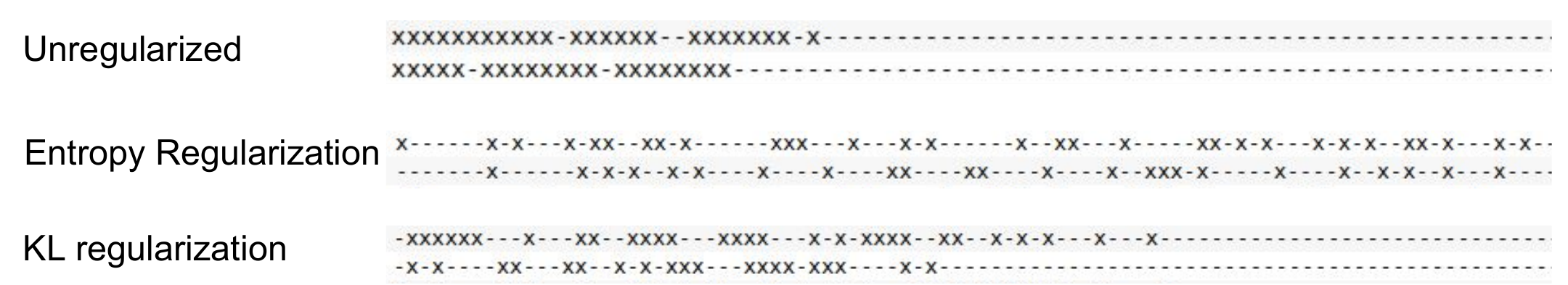}
  \caption{The impact of entropy regularization on emission locations. Each
     line shows the emission predictions made for an example input utterance,
     with each symbol representing 3 input time steps. 'x' indicates that the
     model chooses to emit output at the time steps, whereas '-' indicates
     otherwise. Top line - without entropy penalty the model emits symbols
     either at the start or at the end of the input, and is unable to get
     meaningful gradients to learn a model. Middle line - with entropy
     regularization, the model avoids clustering emission predictions in
     time and learns to spread the emissions meaningfully and learn a model.
     Bottom line - using KL divergence regularization of emission probability
     also mitigates the clustering problem, albeit not as effectively as with
     entropy regularization.}\label{fig:entropy_pen}
\end{figure*}

We now elaborate on the manner in which the gradient is computed.  It
is clear that for a given value of the binary decisions $\tilde{b}_i$,
we can compute $\partial \mathcal{R} /\partial \theta$ using the
backpropagation algorithm.  Figuring out how to learn $\tilde{b}_i$ is
slightly more challenging.  To understand it, we will factor the
reward $\mathcal{R}$ into an expression $\mathcal
{R}(\tilde{\vect{b}})$ and a distribution $\rho(\tilde{\vect{b}})$
over the binary vectors, and derive a gradient estimate with respect
to the parameters of the model:
\begin{equation}
\label{eqn:expectedR}
\mathcal{R} = \E_{\vect{\tilde{b}}} \left[R(\tilde{\vect{b}}) \right]
\end{equation}

Differentiating, we get
\begin{equation}
\label{eqn:grad_reward}
\nabla \mathcal{R} = \E_{\vect{\tilde{b}}} \left[ \nabla R(\tilde{\vect{b}})
       + R(\tilde{\vect{b}}) \nabla \log \rho(\tilde{\vect{b}})
 \right]
\end{equation}
where $\rho(\tilde{\vect{b}})$ is the probability of a binary sequence of the $\tilde{b}_i$
decision variables. In our
model, $\rho(\tilde{\vect{b}})$ is computed using the chain rule over the $b_i$ probabilities:
\begin{equation}
\label{eqn:emit_prob}
\log \rho (\tilde{\vect{b}}) = \sum_{i=1}^T \tilde b_i \log b_i + (1-\tilde b_i) \log (1-b_i)
\end{equation}

Since the gradient in equation~\ref{eqn:grad_reward} is a policy
gradient, it has very high variance, and variance reduction techniques
must be applied. As is common in such problems we use {\it centering}
(also known as baselines) and Rao-Blackwellization to reduce the
variance of such models. See Mnih and Gregor \cite{anvil} for an example of the use of
such techniques in training generative models with stochastic units.

Baselines are commonly used in the reinforcement learning literature to 
reduce the variance of estimators, by relying on the identity
$\E_{\vect{\tilde{b}}}  \left[ \nabla \log \rho(\tilde{\vect{b}}) \right] = 0$. Thus
the gradient in~\ref{eqn:grad_reward} can be better estimated by the following, through the
use of a well chosen {\it baseline} function, $\Omega({\bf x})$, where ${\bf x}$ is a vector
of side information which happens to be the input and all the outputs up to timestep $\tilde{p}_i$:
\begin{equation}
\label{eqn:baselined_reward}
\nabla \mathcal{R} = \E_{\vect{\tilde{b}}} \left[ \nabla R(\tilde{\vect{b}})
       + \left(R(\tilde{\vect{b}}) - \Omega(\bf{x})\right) \nabla \log \rho (\tilde{\vect{b}})
 \right]
\end{equation}
The variance of this estimator itself can be further reduced by Rao-Blackwellization,
giving:
\begin{equation}
\label{eqn:blackwellized_reward}
\E_{\vect{\tilde{b}}} \left[\left(R(\tilde{\vect{b}}) - \Omega(\bf{x})\right)
\nabla \log \rho (\tilde{\vect{b}}) \right] = 
\sum_{j=1}^T \E_{\vect{\tilde{b}}} \left[ \left( \sum_{i=j}^T R_i - \Omega_j \right)
       \nabla \log p (b_t | b_{<t}, \vect{x}_{\leq t}, \vect{y}_{\leq \tilde{p_t}}) \right]
\end{equation}

Finally, we note that reinforcement learning models are often trained with augmented
objectives that add an entropy penalty for actions are the too
confident~\cite{levine2014motor,williams1992simple}. We found this to be crucial for our models to train
successfully. In light of the regularization term, the augmented reward at any time
steps, $i$, is:
\begin{align*}
\label{eqn:entropy_regularization}
	R_i = & \tilde{b}_i \log p(d_i = t_i | \vect{x}_{\leq i},\tilde{\vect{b}}_{<i},
\vect{t}_{<i}) \\ & - \lambda \left(\tilde{b}_i \log p (b_i=1 | b_{<i}, \vect{x}_{\leq i}) 
              +  (1-\tilde{b}_i) \log (p (b_i=0 | b_{<i}, \vect{x}_{\leq i}))\right)
\end{align*}

Without the use of this regularization in the model, the RNN emits all the symbols
clustered in time, either at very start of the input sequence, or at the end. The
model has a difficult time recovering from this configuration, since the gradients
are too noisy and biased. However, with the use of this penalty, the model successfully
navigates away from parameters that lead to very clustered predictions and eventually
learns sensible parameters. An alternative we explored was to use the the KL divergence
of the predictions from a target Bernouilli rate of emission at every step. However,
while this helped the model, it was not as successful as entropy regularization. See
figure~\ref{fig:entropy_pen} for an example of this clustering problem and how
regularization ameliorates it.

\section{Experiments and Results}
\begin{figure}[h]
  \centering
  \includegraphics[width=0.8\linewidth]{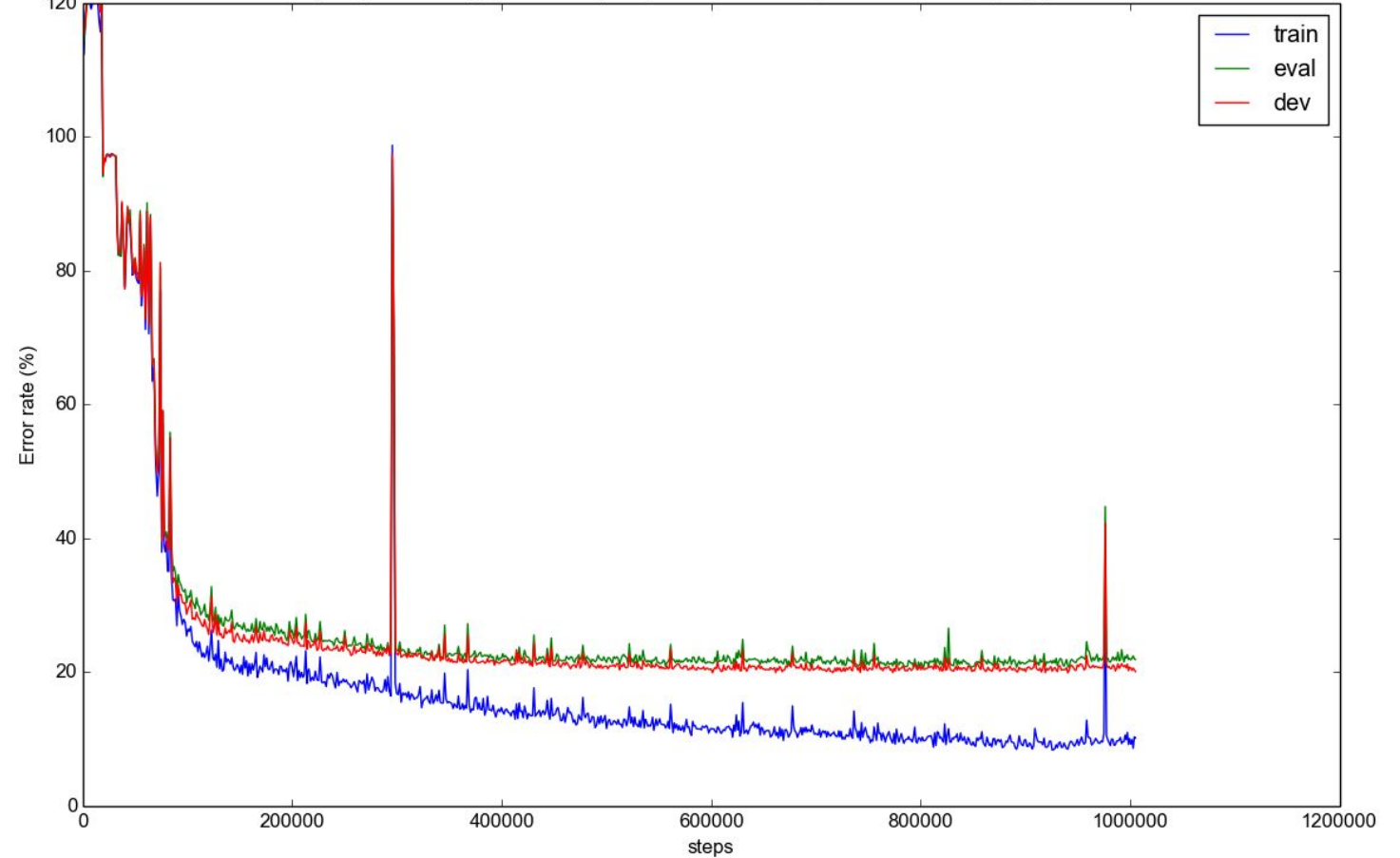}
  \caption{Example training run on TIMIT.}\label{fig:timit}
\end{figure}

We conducted experiments on two different speech corpora using this model. Initial experiments were
conducted on TIMIT to assess hyperparameters that could lead to stable behavior of the model. The
second set of experiments were conducted on the Wall Street Journal corpus to assess if the method worked
on a large vocabulary speech recognition task that is much more realistic and complicated than the
TIMIT phoneme recognition task. While our experiments on TIMIT produced numbers close to the state of
the art, our results on WSJ are only a preliminary demonstation that this method indeed works on such
as task. Further hyperparameter tuning and method development should improve results on this task
significantly.

\subsection{TIMIT}
The TIMIT data set is a phoneme recognition task in which phoneme sequences have to be inferred from
input audio utterances. The training dataset contains 3696 different audio clips and the target is
one of 60 phonemes. Before scoring, these are collapsed to a standard 39 phoneme set, and then the
Levenshtein edit distance is computed to get the phoneme error rate (PER).

We trained models with various number of layers on TIMIT, starting with a small number of layers.
Initially, we achieved 28\% phoneme error rate (PER) using a three layer LSTM model with 300
units in each layer. During these experiments we found that using a weight of $\lambda=1$ for
entropy regularization seemed to produce best results. Further it was crucial to decay this parameter
as learning proceeded to allow the model to sharpen its predictions, once enough learning
signal was available. To do this, the entropy penalty was initialized to 1, and decayed as
$\exp(0.97, \textrm{step}/10000) + 0.1$.  Results were further improved to 23\% with the use of dropout with
15\% of the units being dropped out. Results were improved further when we used five layers of units.
Best results were acheived through the use of Grid LSTMs \cite{kalchbrenner2015grid},
rather than stacked LSTMs.

See figure~\ref{fig:timit} for an example of a training curve. It can be seen that the model
requires a larger number of updates (>100K) before meaningful models are learnt. However, once
learning starts, steady process is achieved, even though the model is trained by policy gradient.

Training of the models was done using Asynchronous Gradient Descent with 20 replicas in
Tensorflow \cite{abadi2016tensorflow}.  Training was much more stable when Adam was used,
compared to SGD, although results were more or less the same when both models were run to
convergence. We used a learning rate of 1e-4 with Adam.  In order to speed up RNN training
we also bucketed examples by length -- each replica used only examples whose length lay within
specific ranges. During training, dropout rate was increased from 0 as the training proceeded.
This is because using dropout early in the training prevented the model from latching on to
any training signal.

Lastly, we note that the input filterbanks were processed such that three continuous frames of
filterbanks, representing a total of 30ms of speech were concatenated and input to the model.
This results in a smaller number of input steps and allows the model to learn hard alignments
much faster than it would otherwise.

Table~\ref{tab:timit} shows a summary of the results achieved on TIMIT by our method and
other, more mature models.
\begin{table}[h]
  \caption{Results on TIMIT using Unidirectional LSTMs for various models.}
  \label{tab:timit}
  \centering
  \begin{tabular}{lll}
    \toprule
    Method     & PER   \\
    \midrule
    Connectionist Temporal Classification (CTC)\cite{graves2013speech}     & 19.6\% \\
    Deep Neural Network - Hidden Markov Model (DNN-HMM)\cite{mohamed2012acoustic}     & 20.7\% \\
    Sequence to Sequence Model With Attention (our implementation)     & 24.5\% \\
    Online Sequence to Sequence Model\cite{jaitly2015online}     & 19.8\% \\
    \midrule
    Our Model (Stacked LSTM)     & 21.5\% \\
    Our Model (Grid LSTM)        & 20.5\% \\
    \bottomrule
  \end{tabular}
\end{table}

\subsection{Wall Street Journal}
\begin{table}[h]
  \caption{Results on WSJ}
  \label{tab:timit}
  \centering
  \begin{tabular}{lll}
    \toprule
    Method     & PER   \\
    \midrule
    Connectionist Temporal Classification (CTC)(4 layer bidirectional LSTM)\cite{graves-icml-2014}     & 27.3 \% \\
    Sequence to Sequence Model With Attention (4 layer bidirectional GRU)\cite{BahdanauCSBB15}    & 18.6\% \\
    \midrule
    Our Model (4 layer unidirectional LSTM)     & 27.0\% \\
    \bottomrule
  \end{tabular}
\end{table}
We used the {\it train\_si284} dataset of the Wall Street Journal (WSJ) corpus for the second
experiment. This dataset consists of more than thirty seven thousand utterances, corresponding
to around 81 hours of audio signals. We trained our model to predict the character seqeunces
directly, without the use of pronounciation dictionaries, or language models, from the audio
signal.  Since WSJ is a larger corpus we used 50 replicas for the AsyncSGD training. Each
utterance was speaker mean centered, as is standard for this dataset.
Similar to the TIMIT setup above, we concatenated three continous frames of filterbanks,
representing a total of 30ms of speech as input to the model at each time step. This is
especially useful for WSJ dataset because its audio clips are typically much longer than
those of TIMIT.
\begin{figure}[h]
  \centering
	\includegraphics[width=0.9\linewidth, trim={2cm 0 2cm 0}]{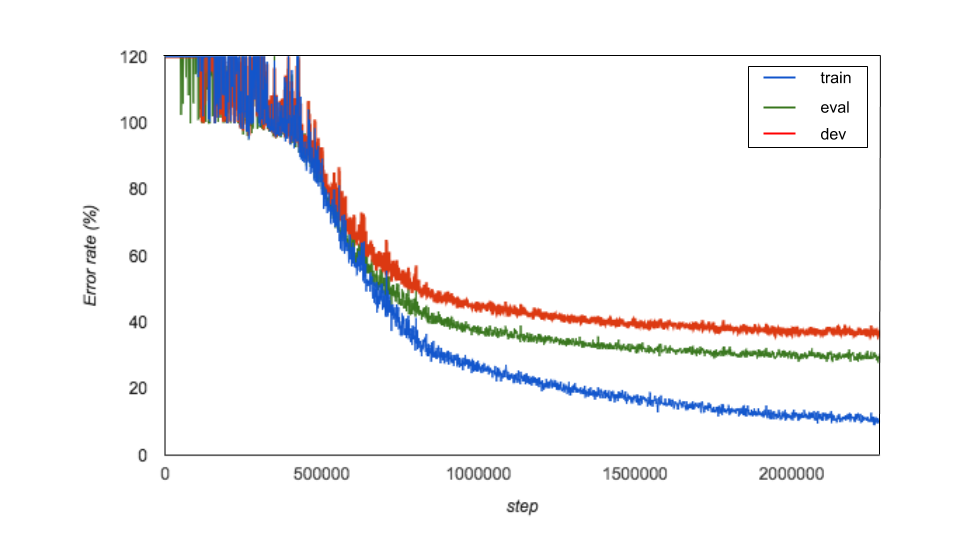}
  \caption{Example training run on WSJ. The pl.}\label{fig:wsj}
\end{figure}

A constant entropy penalty of one was used for the first 200,000 steps, and then it was
decayed as $0.8 * \exp \left(0.97, \textrm{step}/10000 - 20\right) + 0.2$.  Stacked LSTMs with 2
layers of 300 hidden units were used for this experiment\footnote{Admittedly this is a small number
of units and results should improve with the use of a larger model. However, as a proof of
concept it shows that the model can be trained to give reasonable accuracy.}. Gradients
were clipped to a maximum norm of 30 for these experiments.

It was seen that if dropout was used early in the training, the model that was unable to learn.
Thus dropout was used only much later in the training. Other differences from the TIMIT
experiments included the observation that stacked model outperformed the grid LSTM model.
\begin{figure}[h]
  \centering
  \includegraphics[width=1\linewidth,trim=2cm 2cm 2cm 3cm, clip]{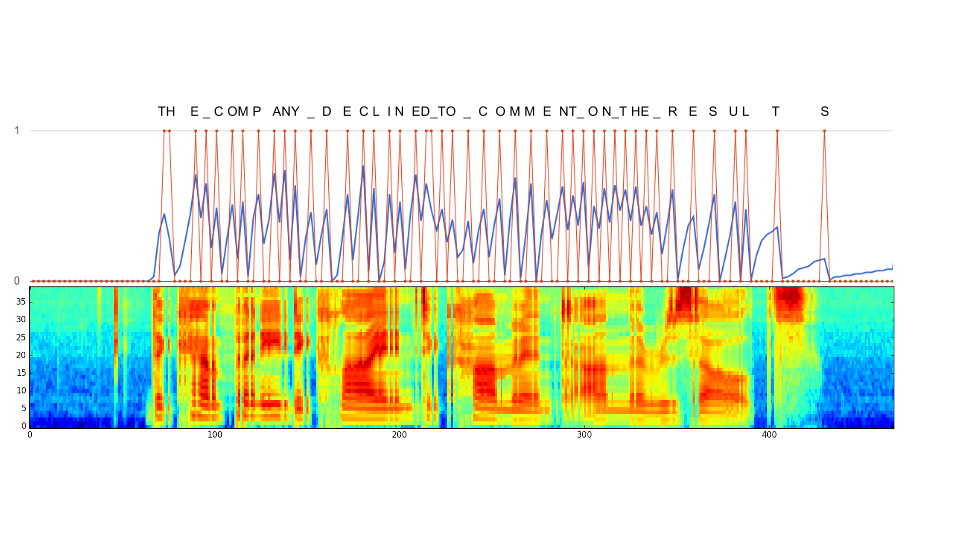}
  \caption{Example output for an utterance in WSJ. The blue line shows the emission
	   probability $\tilde{b}_i$, while the red line shows the
	   discrete emission decisions, $b_i$, over the time steps corresponding to an
	   input utterance. The bottom panel shows the corresponding filterbanks. It
	   can be seen that the model often decides to emit symbols only when new audio
	   comes in. It possibly reflects the realization of the network, that it needs to
	   output symbols that it has heard, to effectively process the new audio.}\label{fig:wsj_emi}
\end{figure}

\subsubsection{Example transcripts}
We show three example transcripts to give a flavour for the kinds of outputs this model
is able to produce. The first one is an example of a transcript that made several errors.
It can be seen however that the outputs have close phonetic similarity to the actual transcript.
The second example is of an utterance that was transcribed almost entirely correctly, other than
the word {\it AND} being substituted by {\it END}. Occasionally the model is even able to transcribe
a full utterance entirely correctly, as in the third example below. 

{\bf REF}: {\it \color{blue} ONE LONGTIME EASTERN PILOT INSISTED THAT THE SAFETY CAMPAIGN INVOLVED NUMEROUS SERIOUS PROBLEMS BUT AFFIRMED THAT
	THE CARDS OFTEN CONTAINED INSUFFICIENT INFORMATION FOR REGULATORS TO ACT ON} \\
{\bf HYP}: {\it \color{red} ONE LONGTIME EASTERN PILOT INSISTED THAT THE SAFETY CAMPAIGN INVOLVED NEW MERCE SERIOUS PROBLEMS BUT AT FIRM THAT
	THE CARDS OFTEN CONTAINED IN SECURITION INFORMATION FOR REGULATORS TO ACT} \\
{\bf REF}: {\it \color{blue} THE COMPANY IS OPENING SEVEN FACTORIES IN ASIA THIS YEAR AND NEXT}\\
{\bf HYP}: {\it \color{red} THE COMPANY IS OPENING SEVEN FACTORIES IN ASIA THIS YEAR END NEXT}\\
{\bf REF}: {\it \color{blue} HE SAID HE AND HIS FATHER J WADE KINCAID WHO IS CHAIRMAN OWN A TOTAL OF ABOUT SIX POINT FOUR PERCENT OF THE COMPANYS COMMON}\\
{\bf HST}: {\it \color{red} HE SAID HE AND HIS FATHER J WADE KINCAID WHO IS CHAIRMAN OWN A TOTAL OF ABOUT SIX POINT FOUR PERCENT OF THE COMPANYS COMMON}\\

\subsubsection{Example Emissions}
Figure~\ref{fig:wsj_emi} shows a plot of the emission probabilities produced as the input
audio is processed. Interestingly, the model produces the final words, only at the end of
the input sequence. Presumably this happens because no new audio comes in after half way
through the utterance, and the model has no need to clear its internal memory to process
new information.


\section{Conclusions}

In this work, we presented a simple model that can solve
sequence-to-sequence problems without the need to process the entire
input sequence first.  Our model directly maximizes the log
probability of the correct answer by combining standard supervised
backpropagation and a policy gradient method.

Despite its simplicity, our model achieved encouraging results on a
small scale and a medium scale speech recognition task.  We hope that
by scaling up the model, it will achieve near state-of-the-art results
on speech recognition and on machine translation, which will in turn
will enable the construction of the universal instantaneous
translator.

Our results also suggest that policy gradient methods are reasonably powerful,
and that they can train highly complex neural networks that learn to make
nontrivial stochastic decisions.

\section{Acknowledgements}
We would like to thank Dieterich Lawson for his helpful comments on the manuscript.

\bibliography{cites}
\bibliographystyle{plain}

\end{document}